\documentclass[lettersize,journal]{IEEEtran}
\usepackage{amsmath,amsfonts}
\usepackage{algorithmic}
\usepackage{algorithm}
\usepackage{array}
\usepackage[caption=false,font=normalsize,labelfont=sf,textfont=sf]{subfig}
\usepackage{textcomp}
\usepackage{stfloats}
\usepackage{url}
\usepackage{verbatim}
\usepackage{graphicx}
\usepackage{cite}
\usepackage{multirow}
\begin{document}

\title{FLORA: Formal Language Model Enables Robust Training-free Zero-shot Object Referring Analysis}

\author{Zhe Chen, ~\IEEEmembership{Member,~IEEE,}, 
Zijing Chen
\thanks{Zhe Chen and Zijing Chen are with Cisco-La Trobe Centre for Artificial Intelligence and Internet of Things, Department of Computer Science and Information Technology, La Trobe University. Edwards Rd, VIC 3552, Australia. Email: \{zhe.chen, zijing.chen\}@latrobe.edu.au
}
}

\markboth{Journal of \LaTeX\ Class Files,~Vol.~14, No.~8, August~2021}%
{Shell \MakeLowercase{\textit{et al.}}: A Sample Article Using IEEEtran.cls for IEEE Journals}


\maketitle

\begin{abstract}
Object Referring Analysis (ORA), commonly known as referring expression comprehension, requires the identification and localization of specific objects within an image based on natural language descriptions. Unlike conventional object detection, ORA requires both accurate language understanding and precise visual localization, making it inherently more complex. Although recent pre-trained large visual grounding detectors have achieved significant progress, these approaches heavily rely on extensively labeled data for fine-tuning and can be burdened by time-consuming learning processes. 
To address these limitations, we introduce a novel, training-free framework for zero-shot ORA, termed FLORA (Formal Language for Object Referring and Analysis). FLORA harnesses the inherent reasoning capabilities of large language models (LLMs) and integrates a formal language model - a logical framework that regulates natural language within structured, rule-based descriptions - to provide effective zero-shot ORA. In particular, with the help of LLMs, our formal language model (FLM) enables an effective, logic-driven interpretation of object descriptions without necessitating any training processes. Built upon FLM-regulated LLM outputs, we further devise a Bayesian inference framework and employ appropriate off-the-shelf interpretive models to finalize the reasoning, delivering favorable robustness against LLM hallucinations and compelling ORA performance in a training-free manner. 
In practice, our FLORA offers significant improvements over existing pretrained grounding detectors, boosting their zero-shot performance by up to around 45\%. Our comprehensive evaluation across different challenging datasets also confirms that FLORA consistently surpasses current state-of-the-art zero-shot methods, providing superior performance in both detection and segmentation tasks associated with zero-shot ORA. We believe our probabilistic parsing and reasoning of the LLM outputs elevate the reliability and interpretability of zero-shot ORA, offering a brand-new solution for ORA tasks. Upon publication, we will make the code available.
\end{abstract}


\section{Introduction}
\IEEEPARstart{R}{eferring} expressions, integral to everyday dialogue and professional interactions, help direct to specific objects within the environment, such as requesting a "cup of tea on the table". Using referring expressions, Object Referring Analysis (ORA) described in this paper seeks to use machines to interpret such natural textual descriptions and accurately identify the corresponding object in an image. In the related field, terms such as referring expression comprehension (REC) \cite{qiao2020referring} and referring expression segmentation (RES) \cite{qiu2019referring} are commonly used. REC mainly focuses on detecting the referred object in the form of bounding boxes, while RES aims at generating segmentation masks. Here, we attempt to unify these terms under the umbrella of the ORA to underscore the overall importance of enhancing language-vision alignment, which could serve as a cornerstone for advancing various AI technologies like visual question answering \cite{wu2017visual, lu2023multi}, text-controlled precise image editing \cite{kawar2023imagic, brooks2023instructpix2pix}, vision-language navigation \cite{wang2021structured, wang2021structured}, and so on

Despite advancements with large vision-language models (LVLMs) like CLIP\cite{radford2021learning, cherti2023reproducible}, ORA remains a challenging task. It demands not only a precise understanding of linguistic semantics but also the capacity for effective visual reasoning, particularly when distinguishing among multiple objects of the same category. Recently, visual grounding detectors \cite{minderer2022simple, minderer2024scaling, liu2023grounding} have shown great potential in addressing ORA by linking textual descriptions with image content during detection. However, without extensive fine-tuning on rich well-annotated ORA datasets like \cite{mao2016generation, yu2016modeling}, state-of-the-art approaches like OWLv2 \cite{minderer2024scaling} and Grounding DINO \cite{liu2023grounding} typically achieve sub-optimal results in ORA. Given greater challenges and higher costs associated with annotating ORA data compared to conventional object detection, the successful development of robust zero-shot ORA techniques thus offers a much more practical and economically viable alternative.

\begin{figure*}[h]
    \centering
    \includegraphics[width=0.88\linewidth, height=0.28\textheight]{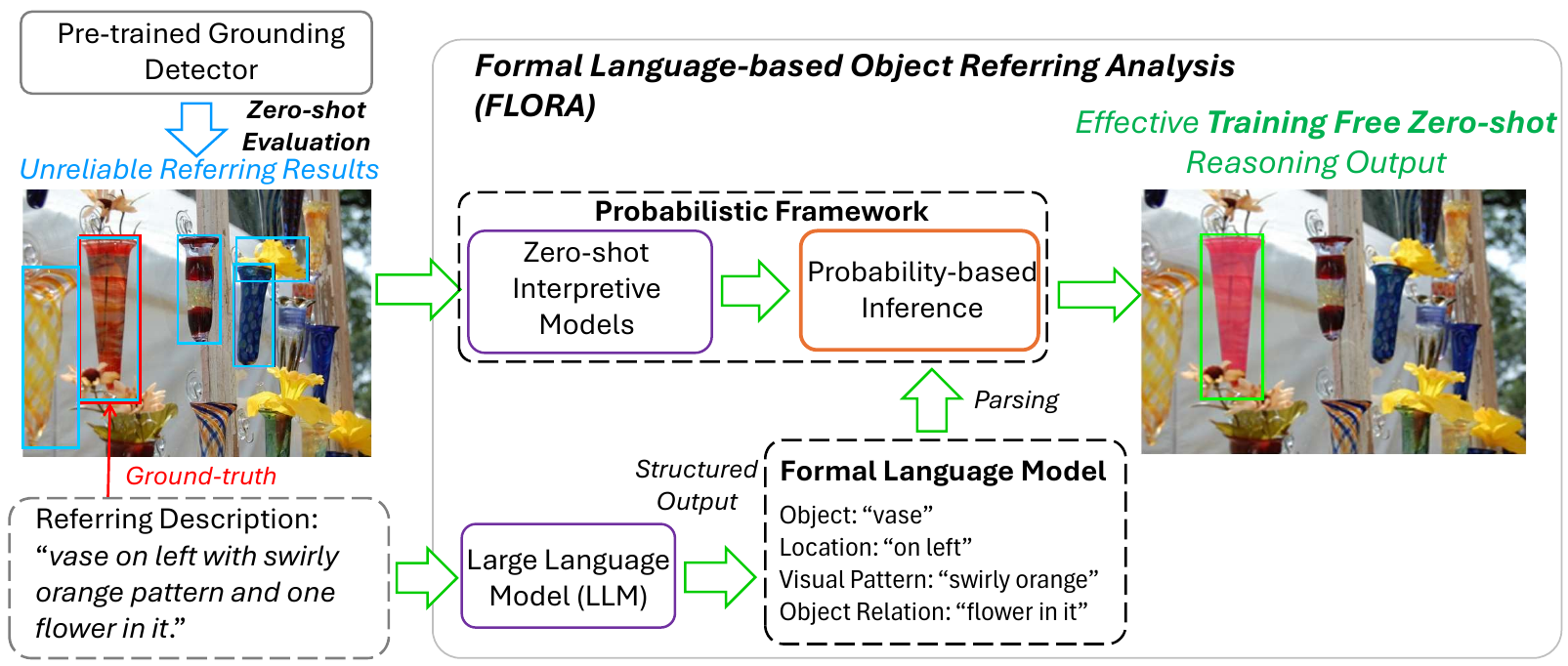}
    \caption{The concept behind our Formal Language-based Object Referring Analysis (FLORA) framework, which enables outstanding accuracy in a training-free zero-shot setting. }
    \label{fig:title}
\end{figure*}

To address the zero-shot ORA, researchers have endeavored to bridge the gap between complex referring phrases and visual images. For instance, some studies \cite{subramanian2022reclip, yu2023zero} employ hand-crafted language parsers to simplify input captions into basic phrases for aiding text-image association. However, these methods often struggle with lengthy, complex captions and fail to grasp nuanced object relationships inherent in ORA. Later, LVLMs that are pre-trained on extensive text-image datasets show great promise due to their generalization capabilities with ORA data \cite{liu2023grounding, wang2022cris}. Yet, they remain susceptible to the complexities of intricate referring phrases and specific object relations. A recent analysis \cite{yuksekgonul2022and} reveals that LVLMs tend to function merely as "bags-of-words," underperforming in tasks that require compositional understanding \cite{doveh2023dense, doveh2023teaching, gupta2023visual, wu2023aligning}. While augmenting prompts \cite{doveh2023dense, doveh2023teaching,yuksekgonul2022and} and enhancing features \cite{jiang2024comclip} have shown potential, the related techniques still face limitations in fully capturing the diversity of referring expressions and can introduce biases detrimental to ORA performance. More recent studies have seen researchers enhance zero-shot capabilities for complex phrase interpretation and nuanced object relation modeling through LLMs. For example, Han et al. \cite{han2024zero} have applied LLMs to translate text and image pairs into object information triplets, enabling their models to pinpoint referred objects based on these triplets effectively. Besides, another study \cite{gupta2023visual} has exploited the in-context learning capabilities of LLMs to produce function-calling parameters that facilitate the extraction of specific object information. Despite these progresses, LLM-based zero-shot ORA is significantly challenged by the possibility of producing hallucinatory content, which can degrade ORA performance. Nevertheless, substantial pre-training remains essential for many of these methods to achieve robust ORA.

Alternative to existing zero-shot ORA methods, we propose that the benefits of pre-trained large foundation models, including both LLMs and LVLMs, can be magnified by ensuring their outputs are well-formulated and can be dynamically parsed. The well-formulated description could simplify the identification of essential object information from the complexities of sentence structures. The dynamic parsing of model outputs, in which object information is probabilistically expressed, could not only deal with complicated descriptions but also mitigate errors such as hallucinations when using LLMs. The organic integration of the well-formulated model outputs and the adaptive probabilistic parsing ensures that LLMs and LVLMs can be appropriately exploited for robust zero-shot ORA without any additional training.

By addressing the zero-shot ORA, we introduce the Formal Language Object Referring Analysis (FLORA) framework consisting of a formal language model for LLMs and a probabilistic framework that helps interpret and reason about LLM outputs, achieving favorable zero-shot ORA performance without relying on any training. 
Specifically, we devise a novel formal language model specifically tailored for the ORA task. The formal language model is a well-developed logical and linguistic framework that formulates complex expressions into a set of syntax and grammar. Fortunately, current LLMs like Llama\cite{touvron2023llama} have already learned formal language knowledge within its model. Without relying on supervised fine-tuning and heavy in-context learning, we found that the formal language knowledge encoded with existing LLMs can directly enable the effective translation of complex referring phrases. Fig. \ref{fig:title} shows an example of utilizing LLMs to generate regulated results for the ORA task, where complex referring phrases can be represented by a few well-formulated components.
After LLM translation, we then parse the LLM-inferred formal languages into object-related semantics. This parsing process assigns precise meanings to these structural outputs.
Afterward, the parsed semantics can then be used to match each potential object found by a grounding detector under a probabilistic framework. That is, based on the Bayesian inference, we feed the parsed semantics into appropriate interpretative models, and the interpretation models will contribute to the calculation of the probabilities that a detected object matches the referring phrase. The obtained probabilities then provide an effective estimation of the referred object, delivering promising \textbf{training-free zero-shot} ORA results. 

Our main contributions to this study can be summarized as follows
\begin{itemize}
    \item We introduce the FLORA, a novel zero-shot ORA paradigm that leverages a formal language model to regulate the outputs of LLMs within a probabilistic parsing framework. FLORA requires no training and demonstrates compelling ORA performance, potentially laying the groundwork for next-generation zero-shot approaches in ORA and other visual reasoning tasks.
    \item We propose the first formal language model that guides LLMs to generate well-structured descriptions. This model operates without training and reduces the need for excessive prompting.
    \item We devise a probabilistic parsing framework that transforms LLM-inferred descriptions into detailed semantics about objects. Utilizing Bayesian inference and tailored interpretive models, this framework assesses the likelihood of each detected object matching the given referring phrase, which effectively fulfills ORA and can tolerate hallucinations from LLMs very well.
    \item We rigorously test our FLORA model across different ORA datasets, demonstrating significant improvements in zero-shot ORA performance over baseline visual grounding models. Our FLORA also achieves state-of-the-art zero-shot ORA accuracy on diverse datasets, showcasing its effectiveness and generalizability across different scenarios. These results validate FLORA’s potential as a versatile tool in multimodal visual understanding and reasoning.
    \item We believe that the FLORA paradigm presents a new foundation for utilizing LLMs and LVLMs in advanced multimodal tasks. This approach holds promise for enhancing data efficiency, reliability, and practicality in high-stakes applications where data collection or annotation is challenging, such as medical diagnosis\cite{litjens2017survey}, human behavior analysis\cite{zhang2021towards}, and autonomous driving\cite{geiger2012we, chen2019progressive}.
\end{itemize}

\section{Related Work}
\subsection{Object Referring Analysis}
Contrary to general image captioning \cite{pan2020x, stefanini2022show, hu2022scaling, li2022comprehending, yao2019hierarchy} that provides discriminative descriptions of objects in an image, ORA aims at locating an object in the image based on referring expressions. In the related field, more commonly used terms for ORA are "referring expression comprehension" (REC) or "referring to image segmentation" (RES). 
Here, we use the ORA to summarize both tasks.  
Up to now, there have been many efforts made to evaluate ORA performance. For example, RefCOCO\cite{yu2016modeling}, RefCOCOg\cite{mao2016generation}, and RefCOCO+\cite{yu2016modeling} are the 3 most popular datasets for benchmarking the REC models, as well as RES models. RefCOCO/RefCOCO+ were built upon the ReferIt Game\cite{kazemzadeh2014referitgame}, and each contains around 50,000 referred objects in around 20,000 images. The RefCOCO/RefCOCO+ uses concise referring expressions, while RefCOCOg uses longer and more complex expressions. Since all these datasets utilize images from the MS COCO dataset \cite{lin2014microsoft}, we can easily obtain the object masks, thus it is easy to extend these datasets into RES datasets. In addition to RefCOCO datasets, there are many similar ORA datasets, such as Whos Waldo\cite{cui2021s}, PhraseCut \cite{wu2020phrasecut}, and so on. However, many of the similar datasets \cite{liu2019clevr, kurita2023refego} are usually designed for some particular scenarios, such as identifying boxes or spheres from the scene \cite{liu2019clevr}, identifying objects with the help of extra scene knowledge \cite{dang2023instructdet}, and so on. Despite these particular settings, REfCOCO-series datasets are still widely accepted for benchmarking ORA models. For more details about ORA tasks, we refer readers to surveys like 
\cite{qiao2020referring}.

Using the available datasets, early fully supervised REC methods \cite{mao2016generation, yu2016modeling, nagaraja2016modeling, hu2016natural, luo2017comprehension, zhang2018grounding} tend to tackle the task using a convolutional neural network (CNN) and recurrent neural network (RNN). If considering different implementation strategies, we have one-stage models \cite{yang2019fast, sadhu2019zero, luo2020multi} and two-stage models \cite{zhang2018grounding, hu2017modeling, wang2019neighbourhood}. Either way, researchers focus on aligning a detected object with the referring expression based on the modeling of the subject, relationship, and location with the help of CNN/RNN. In addition to REC, the RES models \cite{qiu2019referring, luo2020multi} further care about how to generate segmentation outputs by fusing image and text features. Despite promising performance, these methods are sort of limited by the scale of the ORA datasets. In other words, later research found that more text-image data contributes to even higher fully supervised performance. With the help of Transformer-based foundation models like CLIP\cite{radford2021learning}, visual grounding detectors \cite{liu2023grounding} pretrained with a significant volume of text-image data have achieved impressive fully supervised performance. Although LVLMs such as CLIP\cite{radford2021learning, cherti2023reproducible} and FLAVA\cite{singh2022flava} are promising, ORA remains a challenging task. It demands not only a precise understanding of linguistic semantics but also the capacity for effective visual reasoning, particularly when distinguishing among multiple objects of the same category. Recently, visual grounding detectors \cite{minderer2022simple, minderer2024scaling, liu2023grounding} have shown great state-of-the-art performance in addressing ORA by linking textual descriptions with image content during detection. However, without extensive fine-tuning on large well-annotated ORA datasets like \cite{mao2016generation, yu2016modeling}, cutting-edge approaches like OWLv2 \cite{minderer2024scaling} and Grounding DINO \cite{liu2023grounding} typically achieve sub-optimal results in ORA.


\subsection{Grounding Detection}
Grounding detection, which can also be termed open-set object detection or open vocabulary detection, aims at detecting objects based on natural text descriptions \cite{redmon2017yolo9000, li2016attentive, liu2023grounding}. This requires the related model to understand both visual and textual semantics and align them together to deliver the final detection result. Dealing with grounding detection used to be very challenging, since the alignment of multimodal information processing requires complicated design in traditional detection pipelines such as Faster CNN-based \cite{ren2016faster, chen2023transformer} and Yolo-based architectures \cite{redmon2017yolo9000}. Fortunately, with the introduction of Transformer \cite{vaswani2017attention}, the detection-with-Transformer (DETR) architecture \cite{carion2020end} has enabled effective multimodal fusion to facilitate grounding detection \cite{liu2023grounding}. 

Recent research on grounding detection can be roughly categorized into 2 types based on their different focuses. The first type primarily focuses on nouns in the texts and aims to detect objects described by the provided nouns. Representative datasets include Object365 \cite{shao2019objects365} (fixed-category), Flickr30K Entities \cite{plummer2015flickr30k} (open-vocabulary), and so on. 
To achieve this task, OV-DETR \cite{zareian2021open} utilizes a CLIP model to encode images and texts and detect category-specified boxes. Built upon the CLIP, another approach, ViLD \cite{gu2021open}, still employs an R-CNN-like detector to identify the regions containing the semantics of the language.
Besides the nouns, the second type of method generalizes more, aiming to comprehend more complex relations among entities depicted by the texts and provide accurate localization of the specified object. This is more related to the ORA task we study in this paper. 
To tackle this task, there is some noticeable progress, including Grounding DINO \cite{liu2023grounding}, UNINEXT\cite{lin2023uninext}, and so on. These approaches represent state-of-the-art performance in grounding detection. However, they largely rely on large-scale training data.  
Furthermore, although large-scale pertaining or fine-tuning on ORA data can be very effective, these data-demanding models may not necessarily address the reasoning problem explicitly. In fact, it is still unknown to us whether they can generalize well to all real-world scenarios. Recently, some research \cite{mirzadeh2024gsm} even shows negative signs of reasoning using large foundation models. Nevertheless, training these foundational grounding detectors is extremely costly and resource-consuming, especially when ORA annotations are more difficult to acquire.

\subsection{Zero-shot Reasoning for ORA}
Rather than fully supervised grounding detection, zero-shot reasoning on ORA has begun to become a trending topic. This task aims to perform reasoning on ORA without the need for training data from ORA datasets. 
Without using training data on a specific task, we humans could easily transfer our prior world knowledge to unseen scenarios and perform favorably, but this ability is still somewhat beyond the capability of current AI models. Traditionally, to enable zero-shot transfer, especially in natural language processing and computer vision, researchers leverage the pre-trained word embeddings and perform zero-shot predictions by mapping between visual patterns and word embeddings \cite{chen2022transzero, han2021contrastive, kenton2019bert, liu2021isometric}. More recently, the introduction of CLIP \cite{radford2021learning} and ALIGN \cite{jia2021scaling} models opens a new direction - researchers found that, with large-scale image-text pre-training, neural networks could emerge the ability to generalize well on unseen data, showing compelling zero-shot performance in various tasks \cite{baldrati2022effective, zhang2023clamp}. Although not being pretrained on ORA data, the CLIP can still somehow work at associating images with related referring texts, which illustrates significant benefits in zero-shot ORA \cite{hsia2022clipcam}. For example, ViLD \cite{guopen} successfully applied the CLIP model on cropped images for many dense prediction tasks \cite{du2022learning, ghiasi2022scaling}. Adapting CLIP \cite{li2022adapting} took advantage of superpixels to facilitate CLIP to generate high-resolution spatial feature maps and then provide a phrase localization function. A few more related studies \cite{yu2023zero, han2024zero} have been published for zero-shot ORA. For example, Han\emph{et al.} \cite{han2024zero} proposed that zero-shot REC can be fulfilled by refining the modeling of structural similarity between images and texts using CLIP-based encoders. Another study \cite{yu2023zero} then proposed to achieve zero-shot RES by mining better object relations. Despite these progresses, we found that many current zero-shot approaches still somewhat require pre-training to enable their models to generalize well on reasoning, which did not entirely waive the costs of pre-training. Moreover, the obtained zero-shot performance is still suboptimal, leaving a significant gap between the zero-shot performance and the fully-supervised performance.

In addition to current zero-shot ORA approaches, another direction of research aims to equip large foundational models, such as LLMs, with a certain level of reasoning capabilities without training. This research primarily relies on prompting technologies. Thanks to the in-context learning ability of LLMs like Llama\cite{touvron2023llama}, we can use prompts to guide LLMs to generate answers according to specific needs. Current prompting methods include chain-of-thought (CoT) \cite{wei2022chain}, instruction-based methods \cite{instructblip}, compositional reasoning \cite{gupta2023visual}, and so on. More specifically, CoT asks LLMs to perform reasoning by specifying a series of intermediate reasoning steps, which illustrates impressive abilities for solving math and science problems. Different from CoT, instruction-based methods, like InstructBLIP\cite{instructblip}, use a series of instructions as prompts to guide LLMs to generate desired outputs, showing great potential in multimodal reasoning. 
However, the CoT and instruction-based methods are not directly helpful for the zero-shot ORA: the CoT mainly focuses on text-based reasoning rather than visual reasoning, and InstructBLIP still requires pre-training and fine-tuning. Alternatively, a more related strategy is compositional reasoning \cite{gupta2023visual}. This strategy utilizes prompts that help decompose a complex reasoning process into a series of simpler sub-tasks so that specific handlers of different sub-tasks can help finalize the reasoning. The VisProg\cite{gupta2023visual} and Chameleon\cite{lu2024chameleon} characterize the related progress. In these methods, LLMs are prompted to generate structured output in a code style, where the generated "codes" can be directly used to invoke various handlers to fulfill reasoning. 
The handlers may include but are not limited to detectors, segmentation methods, calculation programs, and so on. A major advantage of this is that they are training-free. Using a similar concept, the Langchain \cite{langchain} that formulates LLMs to generate codes for various purposes has shown interesting properties in different applications. However, we would emphasize that these compositional approaches rely heavily on the correct LLM outputs. If the employed LLM has a severe hallucination, the generated "code" would not be useful at all. We usually need a bunch of double-check codes to prevent this, which significantly harms their benefits.

\begin{figure*}[h]
    \centering
    \includegraphics[width=\linewidth, height=0.42\textheight]{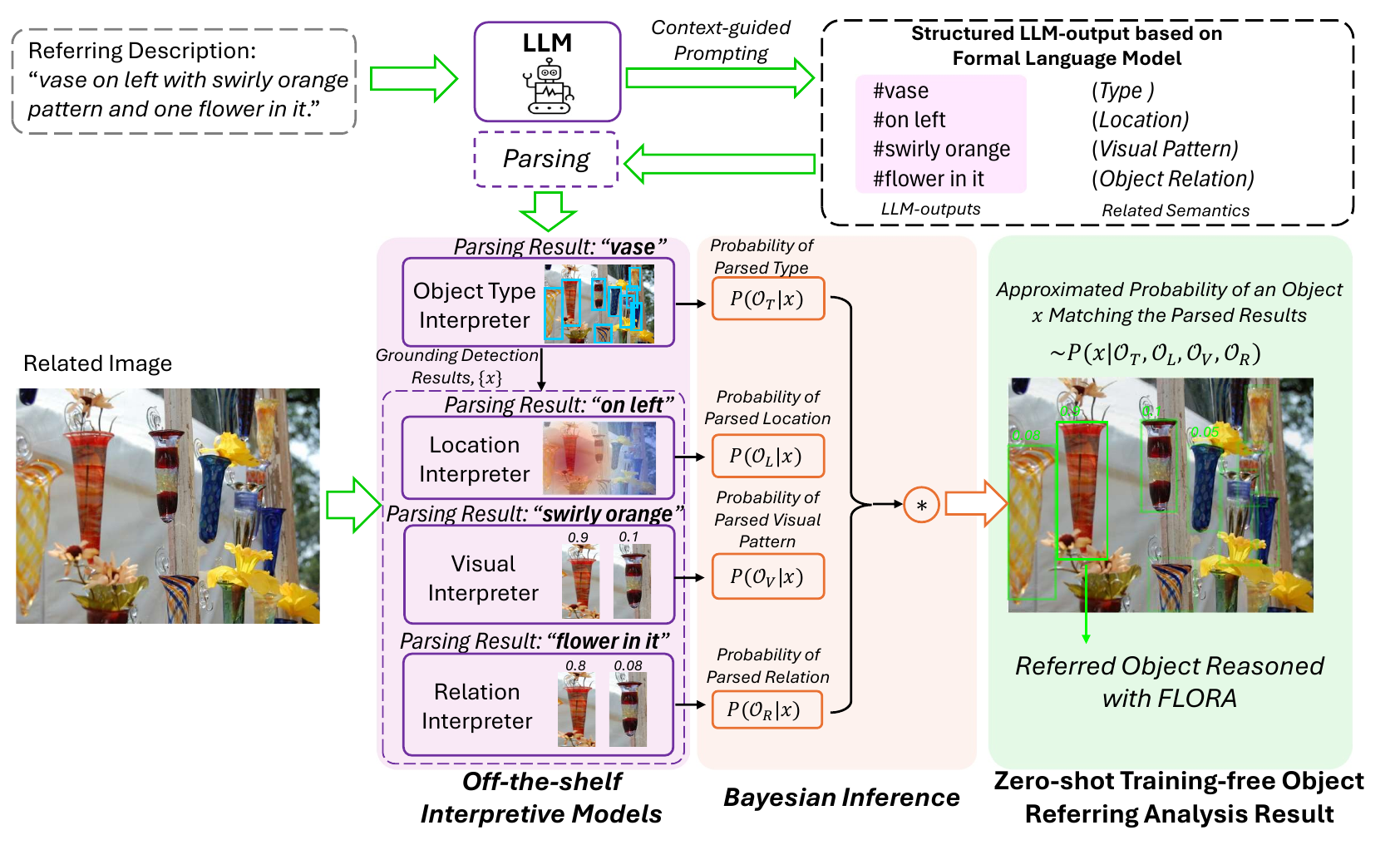}
    \caption{Overall pipeline of our training-free zero-shot FLORA. }
    \label{fig:method}
\end{figure*}

\section{Method}
\subsection{Overview}
Our FLORA methodology mainly consists of a novel formal language model (FLM) and a probabilistic inference framework accompanied by various off-the-shelf interpretative models for \textbf{zero-shot training-free} ORA. Figure. \ref{fig:method} shows the visual summary of the FLORA methodology, outlining the complete workflow from initial input processing to the final object identification. 
More specifically, the FLM is tailored to formulate descriptions of objects from natural language inputs. It directs an LLM, such as Llama\cite{touvron2023llama}, to generate structured outputs that explicitly describe the referred objects in compliance with predefined syntax and semantics. Upon generating the regulated LLM outputs, our FLORA further unfolds in several steps. First, the outputs are parsed to extract detailed information about the object, including its type, location, visual patterns, and relations to other objects. Second, this parsed information is processed through a series of interpreters within a Bayesian probabilistic framework. For example, the object type interpreter can be instantiated by a visual grounding detector like Grounding DINO (GDINO) \cite{liu2023grounding} which identifies objects within the image that align with the prompt described by the parsed object type. 
After interpretive models, by aggregating the obtained probabilities, we assign overall probabilities to detected objects to calculate their overall likelihood of matching the referring phrase. Lastly, the object with the highest matching probability is selected as the one most likely being referred to. This selection process empowers our ORA process to identify and understand referred objects autonomously, without the need for prior training or labeled data.

Based on the  above descriptions, we formally outline our methodology as follows: 
Firstly, we denote a natural referring phrase as $\mathcal{D}$. We utilize an FLM with predefined syntax and semantics to regulate an LLM, enabling it to produce a structured output $\mathcal{S}$ from $\mathcal{D}$: 
\begin{equation}
    \mathcal{S} = F_{LLM}(\mathcal{D}, \mathcal{P}_{FLM}),
    \label{eq:llm}
\end{equation}
where $F_{LLM}$ represents the LLM output guided by the FLM, and $\mathcal{P}_{FLM}$ are the prompts used to regulate the LLM within the FLM. The FLM will be discussed in Sec. \ref{sec:flm} and the prompting will be discussed in Sec. \ref{sec:pmt}. 
Secondly, we parse the structured output $\mathcal{S}$ into specific elements, such as type $S_T$ and spatial location  $S_L$, related to the referred object. We represent the collection of these parsed object semantics as $\mathcal{O} = \{\mathcal{O}_T, \mathcal{O}_L, \ldots\}$:
\begin{equation}
    \mathcal{O} = F_{parse}(\mathcal{S}),
\end{equation}
where $F_{parse}$ is the parsing function, and the details about the parsing will be discussed in sec. \ref{sec:parse}. 
Afterward, using the parsed output $\mathcal{O}$, we employ a probabilistic framework to estimate which detected object best matches the description. Suppose we detected a potential object denoted as $x$, we introduce the probability $P(x|\mathcal{O})$ to quantify how likely it is that $x$ relates to the semantics from $\mathcal{O}$. The details will be presented in Sec. \ref{sec:prob}.
Lastly, we determine the object that is most likely referred to by selecting the one with the highest probability:
\begin{equation}
    \hat{x} = \underset{x\in \mathcal{X}}{\arg\max} P(x|\mathcal{O}),
\end{equation}
where \(\mathcal{X}\) represents the set of all potential objects detected, and \(\hat{x}\) is the object inferred to be the one referred to according to our FLORA framework.

\subsection{Formal Language Model Definition}
\label{sec:flm}
In areas like logic, mathematics, and computer science, formal languages comprise alphabets (symbols) and rules (syntax) for assembling symbols into valid expressions (sentences). For instance, consider a formal language \( L \) defined over an alphabet \( \Sigma = \{a, b, c, d, e, +\} \) with a grammar that includes non-empty strings starting with a "+". In this formalism, strings such as "+abc" and "+de" belong to \( L \), whereas "abb" and "de+" do not. Formal languages are instrumental in formalizing natural languages and defining programming languages, offering a structured approach to manage and reduce output ambiguities. Key benefits of formal languages may include the simplicity of parsing LLM outputs and the lowered risk of hallucinations brought by LLMs. Another benefit could be the fact that cutting-edge LLMs already learn the formal language knowledge in the models. 

For ORA with LLMs, we introduce a tailored FLM to formalize object descriptions derived from natural languages. Our FLM consists of an object alphabet and a set of object syntax rules. The object alphabet defines what information can be used to describe the referred object, and the object syntax defines how this information is supposed to describe the referred object. The detailed definitions of the object alphabet and object grammar are as follows:

\textbf{Object Alphabet:} The alphabet encompasses all symbols utilized by the LLM to generate descriptions, extending beyond English characters to include special symbols like the hashtag (“\#”) which plays an important role in our model.

\textbf{Object Syntax:} The syntax rules of our FLM formulate that valid structured object descriptions commence with a hashtag and unfold as follows: 
\begin{description}
    \item[\texttt{\#[object type]}] \hfill \\
    Specifies the type of the object like 'person' or 'car'. 
    \item[\texttt{\#[spatial location]}] \hfill \\
    Describes where the object is located, \emph{e.g.,} 'left', 'bottom-right', \emph{etc.} 
    \item[\texttt{\#[visual pattern]}] \hfill \\
    Describes visual attributes like 'red', 'shiny', \emph{etc.}
    \item[\texttt{\#[relation to surrounding objects]}] \hfill \\
    Explains how the object relates to others around it, such as 'under the tree', \emph{etc.}
\end{description}
This syntax formulation ensures that each segment of the description is clearly delineated, enhancing the clarity and utility of the original natural language descriptions.
In particular, the use of the hashtag in our model facilitates the segmentation of different descriptive elements, such as object type and spatial location, facilitating the processing and interpretation of LLM outputs.

\textbf{Further Discussion:} While traditional entity identification models statically recognize limited entities (\emph{e.g.}, names, places), they often struggle with the dynamic and varied structure of natural descriptions and are constrained by their training data. In contrast, our FLM offers a greater level of adaptability and does not require retraining to accommodate new types of descriptions by taking the best of existing LLMs using our FLM. Moreover, unlike entity identification which typically ends at recognition, our FLM enhances the understanding of complex relationships between objects and enables more sophisticated text interpretations and reasoning beyond mere entity recognition.

\subsection{Context-guided Prompting for FLM}
\label{sec:pmt}
While our FLM provides a structured approach for LLMs to generate outputs, the LLMs may not inherently adhere to this structure without proper prompting. Thus, we designed an FLM-based context-guided prompting process, termed CoP-FLM (Context-guided Prompting for FLM), to ensure that the LLMs follow the syntax and rules when generating outputs. CoP-FLM integrates FLM-related context and the context information about ORA to guide the LLM, ensuring its response quality. Here, we mainly apply system-level and instance-level context-guided prompting processes. 

At the system level, we employ the following prompt containing FLM-related contexts to guide the LLM:
\begin{quote}
    "\emph{You are an assistant that helps formulate a formal language model. This model describes a referred object based on textual descriptions and a given image. In this model, a referred object is characterized by its object type, spatial location, visual patterns, and its relationships with other objects in the scene. For example, for a description like 'a woman with a red shirt sitting on the bench bottom left,' your responses should be '\#woman' for the object type, '\#bottom left' for the spatial location, '\#wearing a red shirt' for visual patterns, and '\#sitting on the bench' for its relation with other objects.}"
\end{quote}
This prompt broadly outlines the LLM’s tasks, emphasizing the outputs within the confines of our FLM formulation.

At the instance level, we introduce more specific contexts to prompt the LLM to generate outputs that conform to our FLM's object syntax. For instance, to elicit information about the object type, the prompt might be:
\begin{quote}
    "\emph{The description of an object in an image is 'person bottom-left'. Tell me the type of the object described. The answer must start with a \#.}"
\end{quote}
This could lead the LLM to generate a response like '\emph{\# Person}'. Similar prompts are used sequentially to extract details on spatial location, visual patterns, and relations to surrounding objects. Collectively, these responses contribute to a structured composite of object information as prescribed by our FLM.

It is worth mentioning that omitting system-level instructions on using these symbols can lead to inconsistencies in how instance-level prompts are answered, potentially deviating from the desired structured format. Besides, there could be missing information. Regarding this, we prompt the LLM to generate "\# None". 

\subsection{Formal Language Parsing}
\label{sec:parse}

In our FLM, the generated responses are related to object type, spatial location, visual patterns, and relationships to surrounding objects, which are denoted as \(\mathcal{S}_T\), \(\mathcal{S}_L\), \(\mathcal{S}_V\), and \(\mathcal{S}_R\), respectively. These together comprise \(\mathcal{S}\): \(\mathcal{S}=\{\mathcal{S}_T, \mathcal{S}_L, \mathcal{S}_V, \mathcal{S}_R\}\). Now, we define parsing as: 
\begin{equation}
 \mathcal{O} = F_{parse}(\mathcal{S}) = \{V\big(\textit{Spl}(s, \#)\big) | s \in \mathcal{S}\},
 \label{eq:parse}
\end{equation}
where \(\textit{Spl}(\cdot, \#)\) is the function that splits a string according to \#, \(V\) represents a validity filter, and \(s\) is an LLM output from one of the \(\{\mathcal{S}_T, \mathcal{S}_L, \mathcal{S}_V, \mathcal{S}_R\}\). Here, the validity filter helps reduce low-level LLM errors during parsing. It addresses three primary types of inaccuracies in LLM responses: (1) Inaccurate Answers: If the LLM deviates from the FLM format, the responses will be discarded; (2) Redundant Explanations: Excessively verbose explanations are stripped from the output.
(3) Invalid Spatial Information: Common spatial terms can be predefined in a concise dictionary containing terms like "left", "right", and so on. If the parsed location information does not match any entry in this dictionary, we discard the related output. 

The above parsing formula converts raw LLM outputs into structured information, collated into \(\mathcal{O}\).
For example, from a description like "black car on the left", CoP-FLM prompts would guide an LLM to yield "\#Car," "\#Left," and "\#Black" for the type, location, and visual patterns as responses, respectively. These results undergo splitting and validity filtering to produce parsed outputs such as \(\mathcal{O}_T="car"\) for the object type, \(\mathcal{O}_L="left"\) for spatial location, and \(\mathcal{O}_V="black"\) for visual patterns. 

\subsection{Probability Framework for Reasoning and Interpretation}
\label{sec:prob}

\subsubsection{Bayesian Framework for ORA}

Following the parsing of LLM outputs into separate components \(\mathcal{O}_T\), \(\mathcal{O}_L\), \(\mathcal{O}_V\), and \(\mathcal{O}_R\), we utilize the parsed information to facilitate the identification of the referred object. This next step involves interpreting the parsed data using a Bayesian probability framework, which facilitates the zero-shot ORA task and potentially eliminates the need for any training.

Using Bayesian principles, we can formulate the probability that an object \(x\) is the referred object given the structured description \(\mathcal{O}\) as:
\begin{equation}
    P(x|\mathcal{O}) = P(x|\mathcal{O}_T, \mathcal{O}_L, \mathcal{O}_V, \mathcal{O}_R).
\end{equation}
This can be further expressed using Bayesian inference as:
\begin{equation}
    P(x|\mathcal{O}_T, \mathcal{O}_L, \mathcal{O}_V, \mathcal{O}_R) = \frac{P(\mathcal{O}_T, \mathcal{O}_L, \mathcal{O}_V, \mathcal{O}_R|x) \cdot P(x)}{P(\mathcal{O}_T, \mathcal{O}_L, \mathcal{O}_V, \mathcal{O}_R)}.
    \label{eq:prob}
\end{equation}
For better inferring $P(x|\mathcal{O})$ in ORA, we introduce two simplifying assumptions based on above equation: (1) Independence Assumption: The variables \(\mathcal{O}_T\), \(\mathcal{O}_L\), \(\mathcal{O}_V\), and \(\mathcal{O}_R\) are considered independent of each other. We believe this is intuitive. For example, the object type typically does not inherently influence its location in the image.
(2) Uniform Prior Assumption: Both the probability \(P(x)\), representing the likelihood of any object's presence in the image, and \(P(\mathcal{O}_T, \mathcal{O}_L, \mathcal{O}_V, \mathcal{O}_R)\), representing the joint probability of observing the obtained structured outputs, can be assumed uniform and consistent. In other words, we believe it is quite straightforward to assume that the presence of any parsed object characteristics is equally likely, and the overall joint probability does not impact the ORA process.

By applying above assumptions, the Eq. \ref{eq:prob} simplifies to:
\begin{equation}
    P(x|\mathcal{O}_T, \mathcal{O}_L, \mathcal{O}_V, \mathcal{O}_R) \propto P(\mathcal{O}_T|x)P(\mathcal{O}_L|x)P(\mathcal{O}_V|x)P(\mathcal{O}_R|x),
    \label{eq:propto}
\end{equation}
which indicates that determining the $P(x|\mathcal{O})$ breaks down to find the product of the probabilities of each parsed characteristic matching the \(x\). Fortunately, this allows us to leverage various off-the-shelf interpretive models for ORA.

\subsubsection{Probability Interpretive Models}
The Eq. \ref{eq:propto} allows us to leverage pre-trained large foundation models for interpreting object type, location, visual patterns, and relations.

\emph{Object Type Interpreter:}
To interpret and estimate \(P(\mathcal{O}_T|x)\), we use a visual grounding detector like the GDINO\cite{liu2023grounding}  as the model for object type. For instance, if the parsed \(\mathcal{O}_T\) is "person", we utilize "person" as the textual prompt for the grounding detector to detect persons within the image. The confidence scores from these detections provide the probabilities \(P(\mathcal{O}_T|x)\) for each detected object \(x\), formalized as:
\begin{equation}
    P(\mathcal{O}_T|x) = s_T(x),
\end{equation}
where \(s_T(x)\) represents the confidence score for an object \(x\) in the grounding detection results, with \(x \in GroundingDetection(I, \mathcal{O}_T)\) and \(I\) being the input image.

\emph{Spatial Location Interpreter:}  
To compute \( P(\mathcal{O}_L|x) \), we use a location-relevance interpreter, which evaluates the spatial relevance of a detected object \( x \) based on the term \( \mathcal{O}_L \). For instance, if \( \mathcal{O}_L \) is "bottom left," the probability \( P(\mathcal{O}_L|x) \) will be high if \( x \) is positioned with a large horizontal coordinate value and a smaller vertical coordinate value, assuming the image's origin is at the top-left corner. We define this as:
\begin{equation}
    P(\mathcal{O}_L|x) = R(\mathcal{O}_L; L(x)),
\end{equation}
where \( R(\cdot) \) calculates spatial relevance, and \( L(x) \) refers to 
object's spatial information, including horizontal coordinate \( L_h(x) \), vertical coordinate \( L_v(x) \), and size \( L_{sz}(x) \). Suppose \( r_h \), \( r_v \), and \( r_{sz} \) are the relevance scores for horizontal coordinate, vertical coordinate, and size, respectively, then we can have:
\begin{equation}
    R(\mathcal{O}_L; L(x)) = r_h \cdot r_v \cdot r_{sz}.
\end{equation}

To compute relevance, we would like to constrain the \( r_h \), \( r_v \), and \( r_{sz} \) fall within [0, 1] based on a text-instructed spatial interpretation function \( \sigma \). More specifically, the relevance score, \(r\in \{r_h, r_v, r_{sz}\}\) can be computed by:
\begin{equation}
    r = 
    \begin{cases} 
       \sigma(L(x)) & \text{if } t \in \mathcal{O}_L, \\
       \sigma(1 - L(x)) & \text{if } t \notin \mathcal{O}_L,
    \end{cases}
\label{eq:loc}
\end{equation}
where \( t \) is the spatial term in \( \mathcal{O}_L \) related to $r$, and \( \sigma \) can be an absolute, squared, or exponential function. In practice, we found squaring functions are the most effective for ORA tasks.

Implementation of the above equation is simple. 
For example, if \( t \) is a spatial term, \emph{i.e.,} "right", then the horizontal relevance \( r_h = \sigma(L_h(x)) \) if \( \mathcal{O}_L \) contains "right". On the contrary, if it contains "left", then \( r_h = \sigma(1 - L_h(x)) \). Therefore, assuming the origin is at the top-left corner, \( r_h \) increases as \( L_h(x) \) moves toward the right side of the image. Similarly, for vertical relevance \( r_v \), if \(t\) is "bottom" and \(t \in \mathcal{O}_L\), \(r_v\) will increase with larger \( L_v(x) \) values, indicating that objects closer to the bottom of the image are more likely to be referred.

For size relevance \( r_{sz} \), we intuitively infer that closer objects appear larger and further objects appear smaller. As a result, \(t\) would be the term like "close", and the size \( L_{sz}(x) \) is computed as \( \sqrt{height(x) \cdot width(x)} \), where \( height(x) \) and \( width(x) \) are normalized by the height and width of the image. The normalization makes all values in \( L(x) \) fall within the [0, 1] range, ensuring consistent relevance calculation.

\emph{Visual Pattern Interpreter and Object Relation Interpreter:}
Thanks to the recent development of LVLMs like CLIP \cite{radford2021learning} and Grounding DINO \cite{liu2023grounding}\footnote{The grounding detectors can be considered as another type of LVLMs.}, it is now simple to estimate the relevance between a text phrase (\emph{e.g.} the parsed visual patterns \(\mathcal{O}_V\) or object relations \(\mathcal{O}_R\)) and a visual entity (\emph{e.g.} image patch of an object \(x\)). 
Consequently, the probability that the visual patterns of an object \(x\) match the parsed characteristics \(\mathcal{O}_V\), or the probability that the object relations belong to an object \(x\) match the parsed relations \(\mathcal{O}_R\), can be expressed as:
\begin{equation}
    P(\mathcal{O}_V / \mathcal{O}_R|x) = LVLM(str(\mathcal{O}_T,\mathcal{O}_V/\mathcal{O}_R), Region(x, I)),
    \label{eq:vis-rel}
\end{equation}
where \(LVLM\) represents pretrained LVLMs, $str$ is the operation that stacks multiple phrases into a single string-form sentence as the prompt for the VLM, and \(Region(x, I)\) is a function that relates the object \(x\) to a specific region from the input image \(I\) based on the bounding box of \(x\). Here, the slack $/$ means that the interpretations of $\mathcal{O}_V$ and $\mathcal{O}_R$ share a similar formulation. 

When implementing the LVLMs, it is possible to use either the CLIP or the grounding detector like GDINO. CLIP will associate the image patch of \(x\) cropped from image \(I\) with the prompt, and a grounding detector will generate per-prompt relevance estimation \emph{w.r.t.} the detection \(x\). Normally, CLIP is already powerful for this task, but the grounding detector results could be further considered as a complement to the CLIP model. Therefore, we can ensemble the both results to achieve better performance. Empirically, when interpreting visual patterns and object relations contributes 10 points in accuracy (\emph{e.g.} Table \ref{tab:abl-interp}), the CLIP model contributes 9.5 points and the ensemble with grounding detection results contributes another 0.5 points.

As mentioned, when interpreting both visual patterns and object relations, the formulation described in the above equation can be shared, but we would like to emphasize that the prompts are different: interpreting $\mathcal{O}_V$ would involve a prompt focusing the patterns on the target while interpreting $\mathcal{O}_R$ would involve a more complete prompt with a focus on the relations. When running the LVLMs, we actually found that using separate prompts for $\mathcal{O}_V$ and $\mathcal{O}_R$ generates 0.4 point higher accuracy than using a unified single prompt for both  $\mathcal{O}_V$ and $\mathcal{O}_R$. This may partially suggest that our probability decomposition is beneficial. 

\subsubsection{Referring Object Prediction}
After interpreting the parsed terms, we determine the object \(x\) with the highest likelihood of being the referred object. This is achieved by selecting the object with the maximum combined probability from all interpreted categories. The zero-shot ORA task is thus completed as follows:
\begin{equation}
    x^* = \arg\max_x P(\mathcal{O}_T|x)P(\mathcal{O}_L|x)P(\mathcal{O}_V|x)P(\mathcal{O}_R|x).
\end{equation}
This equation finalizes the processing of our FLORA.

Overall, our FLORA operates without the need for any training data or any fine-tuning, yet it achieves promising ORA performance. 
The probability-based interpretation could handle potential inaccuracies or hallucinations from the LLM quite well. If one type of information is (\emph{e.g.} object relation) inappropriately generated by the LLMs due to hallucinations, the other information (\emph{e.g.} \emph{e.g.} visual patterns) could still provide useful indicative results with the help of the interpretive models like the CLIP. 

\section{Experiments}
\subsection{Setup}
To demonstrate the effectiveness of our FLORA, we perform extensive experiments on the popular ORA datasets, such as RefCOCO\cite{yu2016modeling}, RefCOCOg\cite{mao2016generation}, and RefCOCO+\cite{yu2016modeling}. 
Furthermore, we evaluate other datasets to demonstrate that our approach is not subject to the RefCOCO-series datasets. For example, in the referring person detection task, we introduce the WhosWaldo\cite{cui2021s} dataset. In the referring segmentation task, we run experiments on PhraseCut\cite{wu2020phrasecut}. The details of these used datasets are listed as follows:

\subsubsection{\textbf{RefCOCO/RefCOCO+/RefCOCOg}}
The RefCOCO-series datasets collect data from MS-COCO \cite{lin2014microsoft}. They are the most popular datasets for ORA.
All the data in RefCOCO/RefCOCO+/RefCOCOg provide image and expression pairs for representing referring objects. 
RefCOCO has 19,994 images with 142,210 referring phrases. RefCOCO+ has 19,992 images and 141,564 phrases. RefCOCOg contains
26,771 images with 104,560 phrases. It is worth mentioning that phrases in RefCOCO and RefCOCO+, are shorter, averaging 1.6 nouns and 3.6 words. RefCOCOg has longer phrases, averaging 2.8 nouns and 8.4 words. However, the phrases RefCOCO+ exclude the locations to mainly evaluate visual understanding abilities for ORA. All the 3 RefCOCO/RefCOCO+/RefCOCOg datasets provide both detection and segmentation annotations. On these datasets, we follow \cite{wu2020corefqa} and perform similar pre-processing steps.  

On these RefCOCO datasets, evaluation on referring object detection is performed by counting the precision at top 1, 5, and 10 detection results \cite{yu2016modeling, kamath2021mdetr}. The precision is actually described by accuracy, \emph{i.e.}, if the IoU (Intersection over Union) \emph{w.r.t.} the ground-truth box is larger than 0.5, it is a correct detection. Regarding the referring object segmentation task, the performance is estimated based on the IoU between the predicted segmentation mask and the ground-truth mask. The segmentation evaluation is performed using only the top 1 result, but the result of each frame will be aggregated for the whole dataset. That is, an "Overall IoU" and a "Mean IoU" are used for evaluation. The former calculates the sum of all the inter areas divided by the sum of all the union areas for the whole dataset. The latter calculates IoU for each frame and estimates the mean IoU across the whole dataset. For more details, we refer readers to related studies \cite{yu2023zero}.

\subsubsection{\textbf{Who's Waldo}}
Who’s Waldo \cite{cui2021s} introduces a person-centric object-referring task. In this dataset, the caption of an image contains masked names (in the form of "[NAME]") of different persons who appear in the related image. The referring task is performed by linking a specific masked name with the bounding box that covers the corresponding person in the image. For testing, there are 6741 images as the test set. The performance is measured by the accuracy of predicted links \cite{cui2021s}, \emph{i.e.,} the number of correct links against the number of all the links. In this dataset, the detection results of appeared persons are already provided, thus we only need to analyze the caption and make the prediction on the links. More details on adapting our FLORA on this dataset are discussed later.

\subsubsection{\textbf{PhraseCut}}
PhraseCut\cite{wu2020phrasecut} dataset is introduced for referring object segmentation. It contains 77,262 images and 345,486 phrase-region pairs. The dataset generates a challenging set of referring phrases for which the corresponding regions are manually annotated. Phrases of this dataset can correspond to multiple regions. A large number of object and stuff categories are described with their attributes such as color, shape, parts, and relationships with other entities in the image.
When evaluating, researchers also use overall IoU and mean IoU as the metrics. It is worth mentioning that the original PhraseCut paper uses the term "cumulative IoU" as the overall IoU. Here, we still use the term "overall IoU". 

\subsection{Performance on Referring Object Detection}

\subsubsection{\textbf{RefCOCO/RefCOCO+/RefCOCOg}}
\begin{table*}[!t]
\label{tab:refcoco}
\caption{Referring Object Comprehension Performance on RefCOCO/RefCOCO+/RefCOCOg. "ZE" stands for reproduced "Zero-shot Evaluation". The best zero-shot results are highlighted in bold.}
\centering
\resizebox{0.77 \textwidth}{!}{
\begin{tabular}{l |c cc| cc c| cc}
\hline
& \multicolumn{3}{c|}{RefCOCO}  &\multicolumn{3}{c|}{RefCOCO+}  &\multicolumn{2}{c}{RefCOCOg} \\
Method & Val & TestA & TestB & Val& TestA & TestB & Val&Test\\
\hline
Supervised SOTA \cite{lin2023uninext} &  92.6&  94.3 & 91.5 & 85.2 & 89.6&  79.8& 88.7 & 89.4\\
\hline
CPT-Blk w/ VinVL \cite{wu2020corefqa}& 26.9& 27.50& 27.4 &25.4 &25.0 &27.0 &32.1  &32.3 \\
CPT-adapted \cite{subramanian2022reclip} & 23.8& 22.9 &26.0 &23.5& 21.7 &26.3 &21.8 &22.8\\
GradCAM \cite{selvaraju2017grad} &42.9& 51.1& 35.2 &47.8& 56.9 &37.7  &50.9 &49.7\\
ReCLIP \cite{subramanian2022reclip} & 45.8 &46.1 &47.1 &47.9 &50.1 &45.1&59.3 &59.0\\
SS-CLIP\cite{han2024zero} & 60.6& 66.5 &54.9&55.5 &62.6 &45.7&59.9 &59.9 \\
SS-FLAVA\cite{han2024zero}&61.3&60.9 &50.8& 53.4& 47.6& 52.5& 52.7&52.9\\
GVLP \cite{shen2024groundvlp} & 52.6 & 61.3 & 43.5 & 56.4 & 64.8 & 47.4 & 64.3 & 63.5\\
\hline
Owl (ZE baseline)\cite{minderer2024scaling} & 2.4 &	3.5	 &1.9	 &	2.5	 &3.3 &	1.9	 &	3.0	 & 3.2\\
Owl-FLORA (ours) & 33.5	&	38.9&		26.8&			25.0&		32.0&	18.4 &	25.1&	26.0\\
\hline
GDINO (ZE baseline)\cite{liu2023grounding} & 50.4& 57.2 &43.2 & 51.4& 57.6& 45.8 &60.4& 59.5\\
GDINO-FLORA (ours) & \textbf{73.7}	& \textbf{78.5}	& \textbf{67.8}	&\textbf{63.2}&	\textbf{71.6}&	\textbf{53.5}&		\textbf{72.5}	&\textbf{72.1}\\
\hline
\end{tabular}
}
\end{table*}

Firstly, we present the overall comparison of different methods for referring object detection, also known as referring expression comprehension. In this setting, we compare our method against various other well-known and cutting-edge zero-shot referring object comprehension methods, including CPT\cite{wu2020corefqa}, CPT-adapted\cite{subramanian2022reclip}, GradCam\cite{selvaraju2017grad}, ReCLIP\cite{subramanian2022reclip}, REC-CLIP/FLAVA\cite{han2024zero}, and GVLP\cite{shen2024groundvlp}.
The detailed results are shown in Table \ref{tab:refcoco}, where a supervised SOTA method \cite{lin2023uninext} is illustrated for reference. Regarding our methods, we present the performance for using different grounding detectors, \emph{i.e.,} Owl\cite{minderer2024scaling} and GDINO\cite{liu2023grounding}, as the object type interpreter in our FLORA model. The zero-shot evaluation performance of the original grounding detectors is also compared. It is worth mentioning that we do not use the grounding detectors, especially GDINO, pre-trained on RefCOCO datasets to avoid data leak.  

From the results, we can observe that our FLORA using GDINO achieves the best results across all metrics on all the datasets. This demonstrates the superiority of our approach. Although the GDINO itself provides a very good baseline comparable to state-of-the-art zero-shot methods, our FLORA approach further brings up to around 45\% relative improvements, which validates the benefits of our proposed model. For example, on the RefCOCO val, we improve 23.3 points over the GDINO baseline. 
Furthermore, although the Owl detector can hardly detect referred objects, our FLORA can enable the OWL to achieve much more promising ORA results, even comparable to some early zero-shot approaches like CPT-adapted\cite{subramanian2022reclip}. Here, we analyze that the inferior performance of Owl is due to the limit in detection model capacity. Specifically, Owl mainly uses open-vocabulary datasets whose related text prompts are quite simple (\emph{e.g.} a few nouns), while GDINO further uses datasets containing more complex text phrases, making GDINO much more powerful at detecting object types described by various phrases and parsed information. 
Nevertheless, our approach waives any fine-tuning or training, which could make our approach stand out among compared zero-shot approaches like SS-FLAVA\cite{han2024zero} since it still requires learning processes to optimize the object relationships before applying them to the ORA data.

\begin{table}[!t] 
\caption{Referring Object Comprehension on Whos Waldo. \label{tab:waldo}}
\centering
\resizebox{0.62\linewidth}{!}{
\begin{tabular}{l|c}
\hline
Method & Test Accuracy\\
\hline
Supervised\cite{cui2021s} & 63.5 \\
\hline
Random Guessing & 30.7 \\
SL-CCRF \cite{liu2020consnet} & 46.4 \\
SS-CLIP\cite{han2024zero} & 61.3 \\
SS-FLAVA \cite{han2024zero} & 59.8 \\
\hline
FLORA (ours) & \textbf{62.1} \\
\hline
\end{tabular}
}
\end{table}
Regarding the datasets, we would like to emphasize that the RefCOCOg dataset is different from the RefCOCO and RefCOCO+ datasets because the location-related descriptions are intentionally removed from the referring phrases. Without locations, the baseline GDINO method is already very promising by learning from a large amount of visual-text pairs during the pretraining. This is why the improvement of our approach over the GDINO baseline is not as good as the other 2 datasets. Nevertheless, our approach still achieves promising improvements. In particular, when using a weaker grounding detector, \emph{e.g.} OWL, our approach consistently introduces outstanding improvement. To sum up, these results illustrate that location information is very useful in ORA, yet our FLORA still performs well even without locations. 

\textbf{Comparison with training-free methods.} In addition to zero-shot methods, there are a few training-free methods such as Visprog\cite{gupta2023visual} that could be helpful for the ORA task. The Visprog method regulates the LLM to output the code that can be invoked to fulfill specific tasks. 
This method is training-free but requires precise LLM outputs. 
In the experiment, we attempt to compare this training-free method with our approach by adjusting its coding format tailored for the ORA task. For example, we tried to make the LLM output both object detection and location interpretation instructions in its designed code structure. 
However, we simply found that Visprog hardly outputs codes that are effective for ORA. More specifically, we obtain 5.6, 8.5, and 8.8 on RefCOCO Val, TestA, and TestB splits, 4.6, 7.2, and 6.8 on RefCOCO+ Val, TestA, and TestB splits, and 5.5, 8.0 on RefCOCOg Val and Test splits, respectively. 
This shows that Visprog can hardly handle the dynamics in natural referring phrases, only slightly surpassing the zero-shot OWL. We have looked into the running results, we found that there would often be an error in the generated interpretation code when using the Visprog. Sometimes, even the calling function name is wrong.

\subsubsection{\textbf{Whos Waldo}}

\begin{table*}[!t]
 \label{tab:refcoco-seg}
\caption{Referring Object Segmentation Performance on RefCOCO/RefCOCO+/RefCOCOg. "ZE" stands for reproduced "Zero-shot Evaluation". The best zero-shot results in each metric are highlighted in bold.}
\centering
\resizebox{0.85\textwidth}{!}{
\begin{tabular}{c | l |c cc| cc c| cc}
\hline
&& \multicolumn{3}{c|}{RefCOCO}  &\multicolumn{3}{c|}{RefCOCO+}  &\multicolumn{2}{c}{RefCOCOg} \\
Metric&Method & Val & TestA & TestB & Val& TestA & TestB & Val&Test\\
\hline
\multirow{10}{*}{oIoU}&Supervised SOTA \cite{lin2023uninext} (UniNext) &  82.2 & 83.4& 81.3 & 72.5& 76.4& 66.2 & 74.7 & 76.4\\
\cline{2-10}
&Grad-CAM \cite{selvaraju2017grad}  &14.0  & 15.1 &  13.5&  14.5&  15.0&  14.0&  12.5&  12.8 \\
&Mask-CLIP\cite{zhou2022extract} & 19.9 &19.3& 20.2& 20.4 &19.7 &20.8& 18.9& 19.2 \\
&Adapt-CLIP\cite{li2022adapting}&21.7 &20.3 &22.6& 22.6& 20.9& 23.5& 25.5& 25.4\\
&Cropping + CLIP \cite{yu2023zero}&22.7 &22.7& 21.1 &23.1& 22.4 &22.4& 28.7 &27.5\\
&GL-CLIP \cite{yu2023zero} &24.9 &23.6& 24.7& 26.2& 24.9& 25.8 &31.1 &30.9\\
&TAS \cite{suo2023text} &29.7 &32.8& 28.2& 33.5& 40.6& 27.7 &36.8 &37.1\\
\cline{2-10}
&OWL\cite{minderer2024scaling} + SAM \cite{kirillov2023segment} (ZE baseline) & 2.2	&2.9&	2.7	&	2.2&	2.6	&2.5	&2.7&2.8\\
&OWL-FLORA (ours) &20.2	&20.8	&20.8	&	15.9&	18.1	&15.0&	20.2&	17.8\\ 
\cline{2-10}
&GDINO\cite{liu2023grounding}+ SAM\cite{kirillov2023segment} (ZE baseline) &35.3&41.7&30.9	&37.4&43.5&32.4	&41.0& 41.3	\\
&GDINO-FLORA + SAM (ours) & \textbf{55.7}& \textbf{61.5}& \textbf{48.6}	& \textbf{45.8}& \textbf{52.9}	& \textbf{37.2}	& \textbf{51.7}& \textbf{51.8} \\
\hline
\hline
\multirow{9}{*}{mIoU}&Grad-CAM \cite{selvaraju2017grad}  &14.2 &15.9 &13.2 &14.8& 15.9& 13.8 &12.5& 13.2 \\
&Mask-CLIP\cite{zhou2022extract} & 21.3& 20.9& 21.6& 21.6& 21.2 &22.3 &20.1 &20.4  \\
&Adapt-CLIP\cite{li2022adapting}&23.4 &22.1 &24.6& 24.5&22.6& 25.4& 27.6& 27.3 \\
&Cropping + CLIP\cite{yu2023zero}& 24.8& 22.6& 25.7& 26.3 &24.1& 26.5 &31.9& 30.9\\
&GL-CLIP \cite{yu2023zero} &26.2& 24.9 &26.6& 27.8& 25.6& 27.8& 33.5& 33.7\\
&TAS \cite{suo2023text} &39.9 &42.8& 35.8& 43.9& 50.5& 36.4 &47.6&47.4\\
\cline{2-10}
&OWL\cite{minderer2024scaling} + SAM \cite{kirillov2023segment} (ZE baseline)  &2.4&	2.8&	2.6	&	2.4	&2.7	&2.5	&	3.1&	3.0\\
&OWL-FLORA + SAM (ours) &20.8&	20.8&20.3	&16.1&17.7&15.3	&20.8&18.3\\
\cline{2-10}
&GDINO\cite{liu2023grounding} + SAM \cite{kirillov2023segment} (ZE baseline) &44.3	&50.7	&38.7	&	45.8&	51.1&	40.4	&	50.4	&50.4	\\									
&GDINO-FLORA + SAM (ours) & \textbf{64.0}& \textbf{68.1}& \textbf{57.8}	& \textbf{55.2}& \textbf{62.2}	& \textbf{46.2}	& \textbf{61.1}& \textbf{60.7} \\
\hline
\end{tabular}
}
\end{table*}

In addition to RefCOCO-series datasets, we further evaluate our method on the Whos Waldo. Before introducing the detailed results, we would like to discuss the modification in implementing our approach on this dataset. In general, the reason why we need the modification is because this dataset only cares about the link between a name tag in the referring phrase and the related person in the image. The detection results are already provided and we only need to fulfill the association task. For more details about the task, we refer readers to the original paper \cite{cui2021s}. As a result, using this dataset can remove the necessity of a grounding detector in our approach. We then reformulate our FLM to only care about the potential location, visual patterns, and object relations in the image that can relate a detected person to a specific name tag. By performing this modification, we obtain the results as shown in Table \ref{tab:waldo}. 
In the results, we can observe that the cutting-edge zero-shot method SS (SS-CLIP)\cite{han2024zero} has a very promising performance, only slightly worse than the supervised method. When using our approach, we further improve over the SS-CLIP and take a step closer to the supervised method. This demonstrates that our approach of performing reasoning is consistently effective across different referring object comprehension datasets. 

\subsection{Performance on Referring Object Segmentation}

\subsubsection{\textbf{RefCOCO/RefCOCO+/RefCOCOg}}
In this section, we present the overall comparison of different methods for referring object segmentation. In this setting, we compare our method against other well-known and cutting-edge zero-shot referring object segmentation methods, including Grad-CAM\cite{selvaraju2017grad}, Mask-CLIP\cite{zhou2022extract}, Adapt-CLIP\cite{li2022adapting}, Cropping + CLIP\cite{yu2023zero}, GL-CLIP\cite{yu2023zero}, and TAS\cite{suo2023text}. 
The detailed results are shown in Table \ref{tab:refcoco-seg}. As mentioned previously, the referring object segmentation mainly reports the overall IoU, termed oIoU, and mean IoU, termed mIoU. Fortunately, with the introduction of the SAM model which can segment almost everything from the image given prompting bounding boxes \cite{kirillov2023segment}, our FLORA approach can fulfill the segmentation without training. 
Similar to the detection task, we present the performance for using different grounding detectors, \emph{i.e.,} Owl\cite{minderer2024scaling} and GDINO\cite{liu2023grounding}, as the object type interpreter in our FLORA model. The zero-shot evaluation performance of the original grounding detectors is also compared. 

The results continue to show that our FLORA using GDINO has the best accuracy scores on the segmentation across different metrics on all the RefCOCO/RefCOCO+/RefCOCOg datasets. Since we only apply the SAM on top of the baseline methods and our corresponding FLORA methods, the trends of improvements are similar to those in the referring object detection evaluation. Briefly speaking, the GDINO provides good baseline performance, and our method achieves outstanding improvement over the GDINO. It is worth highlighting that our approach greatly surpasses the existing zero-shot referring object segmentation methods such as GL-CLIP\cite{yu2023zero}.

\subsubsection{\textbf{PhraseCut}}
\begin{table}[!t]
\caption{Referring Object Detection on PhraseCut}
\label{tab:phrase}
\centering
\begin{tabular}{l|c|c}
\hline
Method & Accuracy (oIoU) & Accuracy (mIoU) \\
\hline
Supervised\cite{zhang2024groundhog} & 54.5 & -\\
\hline
GL-CLIP\cite{yu2023zero} & 23.6 & - \\
TAS\cite{suo2023text} & 25.6 & 24.6\\
\hline
FLORA (ours) & \textbf{47.0} & \textbf{48.7} \\
\hline
\end{tabular}
\end{table}

For referring object segmentation, there is another popular dataset, PhraseCut, in addition to RefCOCO-series datasets. 
Since it mainly reports against oIoU, we also report the oIoU in this evaluation. The detailed results are shown in Table \ref{tab:phrase}. 
The presented results illustrate that our approach significantly outperforms the current stage-of-the-art zero-shot referring object segmentation method, GL-CLIP\cite{yu2023zero}, which validates that our approach can generalize well to different data. Compared to the supervised cutting-edge performance, our approach is only 8.6 \% short in accuracy. Considering that we do not need any training procedures, this illustrates that our approach presents an effective and efficient usage of various interpretive AI models for reasoning about referring objects.

\subsection{Ablation Studies}
In this section, we demonstrate the effectiveness of the different components of our FLORA. We mainly use the RefCOCO dataset for evaluation due to its comprehensiveness and popularity. 
In our ablation studies, we first study the effect of regularizing LLM outputs with our proposed formal language model. Then, we test the benefits of different components in the context-guided prompting procedure.  Afterward, we compare the benefits of different parsed information. Starting with object type (visual grounding detection), we introduce the interpreted probabilities of location, visual patterns, and object relations, one after another, to observe the differences in improvement.
Lastly, we evaluate the effects of different implementation choices of the employed LLMs.

\begin{table}[!t]
\label{tab:abl-flm}
\caption{Effect of the Formal Language Model and Related Prompting on RefCOCO. Parsing and interpretation are included.}
\centering
\begin{tabular}{l|c |c |c }
\hline
 Method & Val & TestA & TestB \\
\hline
GDINO \cite{liu2023grounding} (ZE baseline) & 50.4 & 57.2 & 43.2\\
\hline
prompt w/o FLM  & 59.6 & 63.9 & 54.5\\
prompt w/ FLM, w/o System-level Prompt & 67.3 & 72.5 & 60.7 \\
prompt w/ FLM, w/ System-level Prompt (ours) & 73.7 & 78.1 & 67.8 \\
\hline
\end{tabular}
\end{table}

\textbf{Formal language model and Prompting:} 
As discussed previously, our FLM could be more resistant to hallucinations with the help of a probabilistic interpretation framework. Such adaptivity is specifically beneficial for the ORA task, since the complexity in natural languages may likely make LLMs generate ambiguous responses. To illustrate such benefits, we compare the effects of using our FLM with the effects of not using our FLM for translating referring phrases in the ORA task. For the one \emph{without} our FLM, there will not be syntax and grammar requirements when prompting. To facilitate the extraction of information that is similar to FLM, we attempted to ask LLM to generate direct answers on the object type, location, visual patterns, and object relations, subsequently, and we extract useful information based on the new line symbol "\textbackslash n". The extracted information will still be compatible with our parsing and interpretation steps. Table \ref{tab:abl-flm} shows the results. 

Firstly, the results illustrate that our parsing and interpretation are already promising without FLM, but the improvement is very limited. Instead, the FLM significantly boosts the improvement. In addition, we also prove that the system-level prompting of our CoP-FLM process is important. It generates very promising improvement, which suggests that in-context learning is a necessary step to make the LLM acknowledge how to answer according to the FLM.  
Lastly, despite the more conservative improvement, the favorable enhancement from our parsing and interpretation steps even without FLM suggests that our proposed probability-based framework is very robust against various LLM outputs and can offer promising improvement on zero-shot ORA. 

\begin{table}[!t]
\label{tab:abl-interp}
\caption{Effect of Different Interpretation Components on RefCOCO.}
\centering
\begin{tabular}{l|c |c |c }
\hline
 Method & Val & TestA & TestB \\
\hline
GDINO \cite{liu2023grounding} (ZE baseline, also Object Type) & 50.4 & 57.2 & 43.2\\
\hline
~~~ + Spatial Location  & 62.7 & 67.4 & 58.1 \\
~~~ + Visual Patterns & 70.5 & 76.5 &64.5\\
~~~ + Object Relation  & 73.7	&78.5	&67.8 \\
\hline
\end{tabular}
\end{table}

\textbf{Interpretative Models:} Our FLORA approach parses the FLM-regulated LLM outputs into several components, including object type, spatial location, visual patterns, and object relations.
In this experiment, we test the importance of each parsed component for contributing to the final zero-shot ORA performance. It is worth mentioning that since we directly employ the grounding detector as our object type interpreter, we mainly test the improvements brought by our location interpreter, visual interpreter, and object relation interpreter. Table \ref{tab:abl-interp} shows the detailed results. Based on the results, it illustrate that the location interpreter contributes significantly to improving the accuracy of zero-shot ORA. For example, the location interpreter improves more than 20\% relative to the GDINO baseline. The visual interpreter, which is the use of CLIP in our implementation, provides a further 10\% relative improvement compared to the baseline. Regarding the object relation interpreter, it provides conservative improvement. We analyze that this is mainly because the spatial location and visual patterns are already very informative. Although the improvement is smaller, the object relation interpretation is still beneficial. 

\begin{figure}[h]
    \centering
    \includegraphics[width=\linewidth, height=0.3\textheight]{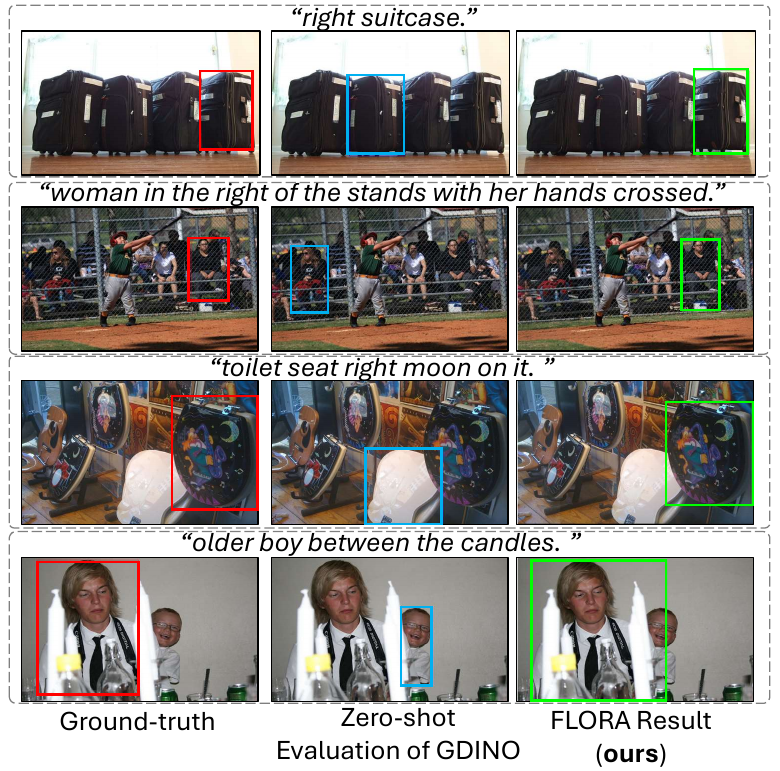}
    \caption{Some visualized results of ORA using GDINO\cite{liu2023grounding} baseline and our FLORA.}
    \label{fig:qua}
    \vspace{-0.4cm}
\end{figure}

\textbf{Additional Study on Interpretive Models:} (1) 
For the spatial location interpreter, as described by Eq. \ref{eq:loc}, there is a function \(\sigma\) that determines relevance scores based on alignment between object locations and parsed information. The \(\sigma\) could be in the form of linear, squared, cubic, and exponential functions. We tested the effects of these functions and obtained an accuracy of 68.7, 70.7, and 66.9 on the Val split of RefCOCO, respectively. This illustrates that the squared function provides a better improvement, but others are also beneficial as long as the function is monotone according to the spatial indicator. If using cubic, however, the weighting would be too much, generating inferior results. (2) We also studied the differences in using various CLIP models, such as ViT-L/H/G versions. We found that the ViT-H is the best but others do not vary too much, with only a 0.5 to 0.8 point drop on the RefCOCO val split.

\textbf{Large Language Models:} We further carry out detailed studies about the impacts brought by using different LLMs. Our experiment is mainly based on the open-source LLM implementation\cite{touvron2023llama}, \emph{e.g.} Llama2 and Llama3, for analysis. We chose these Llama-series LLMs because they are powerful and can be easily set up for use. We believe other LLMs are also useful, but the studies of LLMs themselves are beyond the scope of this study. Here, we only present the differences of using different Llama models in Table \ref{tab:abl-llm} as a brief demonstration. From the results, we can observe that the difference in performance is marginal when using different LLMs, while the Llama3 is slightly better. Since Llama3 is more powerful than Llama2 models, it indeed provides better referring phrase understanding capability. Using Llama2, we have compared a 13b version and a 7b version. Despite more parameters, the Llama2-13b version does not show superiority over the 7b version. Overall, these results illustrate that better-performing LLMs might help improve the referring phrase understanding, but the improvement would not be outstanding. 

\begin{table}[!t]
\label{tab:abl-llm}
\caption{Effect of LLMs for FLORA on RefCOCO.}
\centering
\begin{tabular}{l|c |c |c }
\hline
 Method & Val & TestA & TestB \\
\hline
GDINO \cite{liu2023grounding} (ZE baseline) & 50.4 & 57.2 & 43.2\\
\hline
FLORA - Llama2-7b   & 73.4 &78.4 & 65.8 \\
FLORA - Llama2-13b  & 73.2& 78.0 & 66.3 \\
FLORA - Llama3-7b & 73.7	&78.5	&67.8 \\
\hline
\end{tabular}
\end{table}

\subsection{Qualitative Analysis}
We present some visualized results in Fig. \ref{fig:qua} to better illustrate our benefits. We compare our result with the baseline GDINO \cite{liu2023grounding} approach under zero-shot evaluation setting. From the figures, we can first observe that our approach is better at understanding spatial location information (\emph{e.g.} figures of the 1st and 2nd rows). Besides, our FLORA also significantly enhances the visual and object relation interpretation capability of the baseline GDINO (\emph{e.g.} figures of the 3rd and 4th rows). On one hand, our FLORA integrates interpretive models to facilitate understanding. On the other hand, our FLM makes it possible to translate natural descriptions into much more appropriate prompts for our employed interpretive models using LLMs, and the probability framework organically connects the parsed LLM outputs with the interpretive models.

\section{Conclusion}
In this work, we introduced FLORA, a novel zero-shot, training-free framework that leverages large language models (LLMs), a formal language model (FLM), and various interpretive models integrated with probabilistic reasoning for effective Object Referring Analysis (ORA). 
In particular, by formulating natural language descriptions into structured representations, the FLORA enhances interpretability and mitigates issues like hallucinations inherent in LLMs. Our probabilistic framework bridges the gap between textual semantics and visual grounding, enabling a seamless and robust zero-shot ORA pipeline. Extensive evaluations on multiple ORA datasets, including RefCOCO, RefCOCO+, RefCOCOg, and PhraseCut, demonstrate that our FLORA consistently achieves state-of-the-art zero-shot performance, outperforming existing methods by significant margins. Notably, FLORA improves upon visual grounding detectors by up to 45\% in zero-shot settings and exhibits outstanding generalization across detection and segmentation tasks. These results highlight FLORA's capability to operate effectively without additional training or fine-tuning, presenting a paradigm shift towards data-efficient and scalable ORA solutions.

FLORA's modular design and strategized utilization on off-the-shelf interpretive models make it adaptable to diverse multimodal tasks, including bioinformatics, autonomous navigation, and human behavior analysis. Future research can explore the expansion of FLORA to other vision-language and multimodal reasoning domains and its integration with emerging large foundation models to further enhance zero-shot capabilities. By addressing the challenges of label scarcity and computational overhead, FLORA sets a new benchmark for training-free visual reasoning, paving the way for more interpretable and practical AI systems.

\bibliographystyle{IEEEtran}
\bibliography{ref}

\newpage

\section{Biography Section}
 
\vspace{11pt}

\vspace{-33pt}
\begin{IEEEbiography}[{\includegraphics[width=1in,height=1.25in,clip,keepaspectratio]{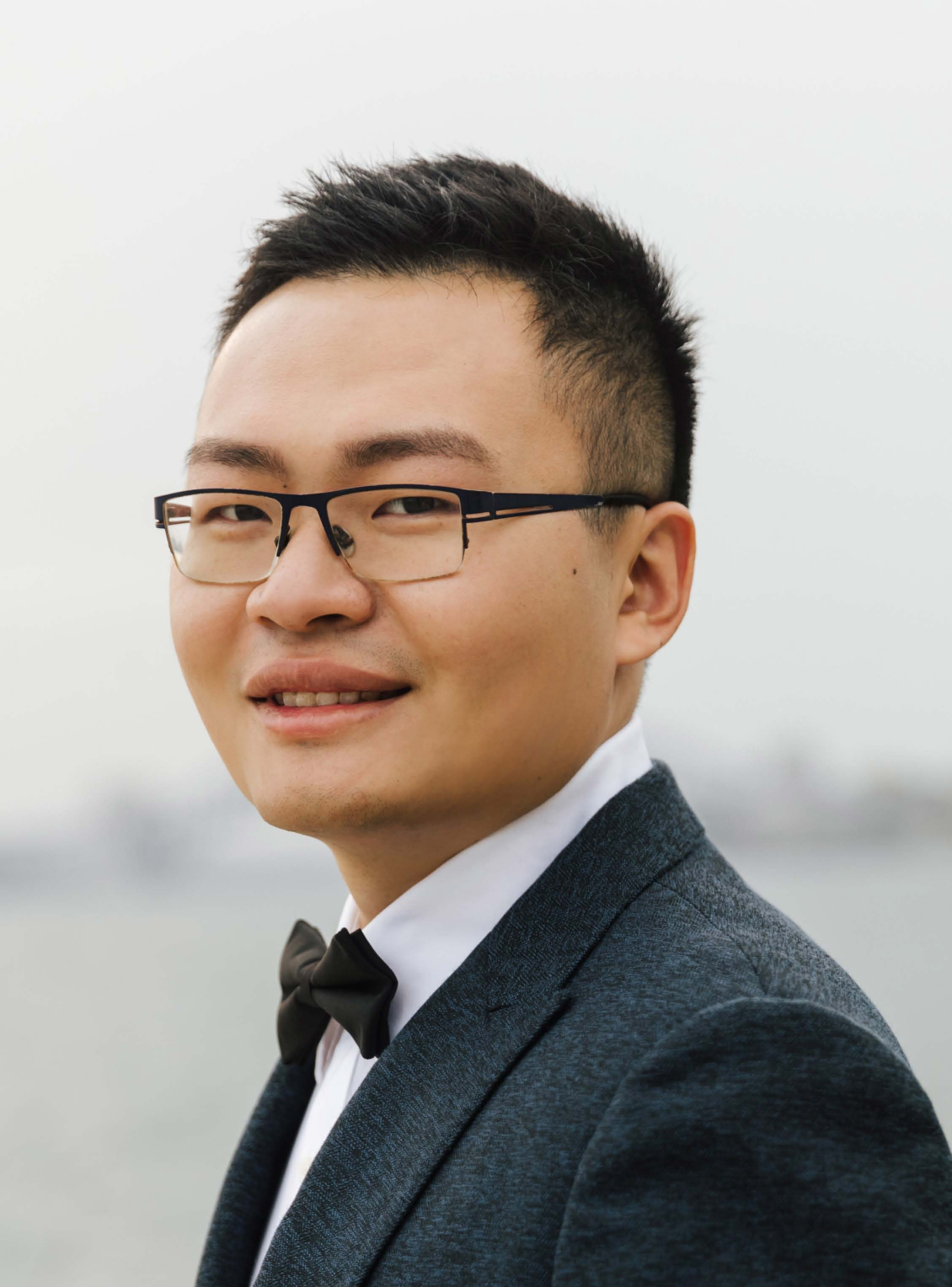}}]{Dr. Zhe Chen}
received his PhD from the University of Sydney in 2019. He is currently a lecturer at La Trobe University and is affiliated with the Cisco-La Trobe Centre for Artificial Intelligence and Internet of Things, as well as the Australian Centre for Artificial Intelligence in Medical Innovation (ACAMI). He is a well-cited researcher, with publications in top-tier venues such as CVPR, ECCV, ICCV, IJCV, and TIP. He has participated in and performed favorably in major international computer vision competitions, including winning a championship at ImageNet. His research focuses on advancing visual understanding and reasoning, with applications spanning healthcare, robotics, and space exploration, among other areas
\end{IEEEbiography}

\vspace{-33pt}
\begin{IEEEbiography}[{\includegraphics[width=1in,height=1.25in,clip,keepaspectratio]{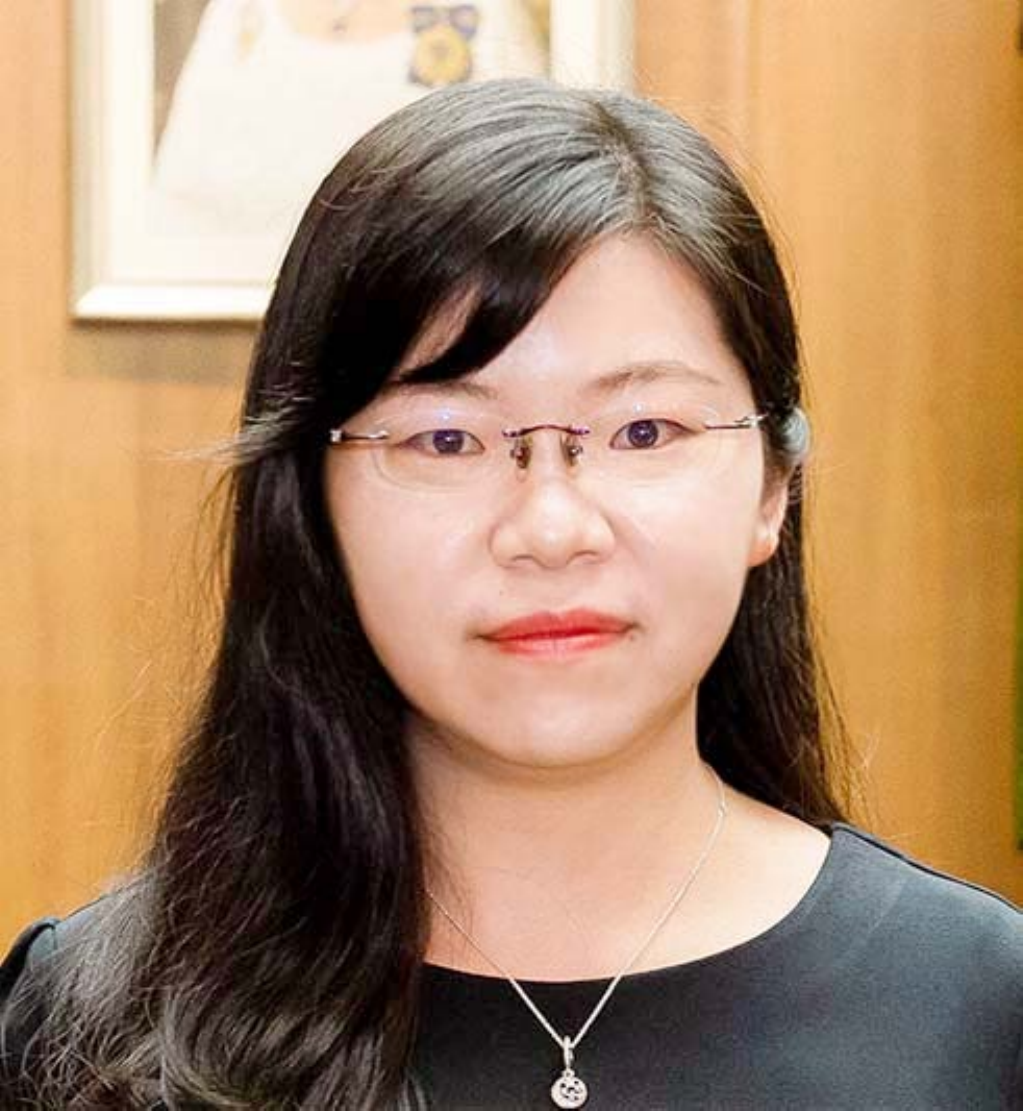}}]{Dr. Zijing Chen} earned her Ph.D. in 2019 from the University of Technology Sydney, specializing in video understanding and processing. She is now working as lecturer at La Trobe University. 
Her research focuses on computer vision and artificial intelligence, including visual object tracking, video processing, scene understanding, and multimodal AI. Her work has been published in leading journals and conferences.
\end{IEEEbiography}
\vspace{11pt}

\vfill

\end{document}